\documentclass[10pt, a4paper]{article}
\usepackage{lrec}
\usepackage{multibib}
\newcites{languageresource}{Language Resources}
\usepackage{graphicx}
\usepackage[dvipsnames]{xcolor}
\usepackage{tabularx}
\usepackage{booktabs}
\usepackage{soul}

\usepackage{epstopdf}
\usepackage[OT1]{fontenc}
\usepackage[utf8]{inputenc}

\usepackage{hyperref}
\usepackage{xstring}

\newcommand{\embosi}{Mboshi}
\newcommand{\ligaikuma}{Lig\_Aikuma}
\newcommand{\dpseg}{dpseg}

\makeatletter
\newcommand\footnoteref[1]{\protected@xdef\@thefnmark{\ref{#1}}\@footnotemark}
\makeatother

\usepackage{newunicodechar}
\usepackage{tipa}
\usepackage{ upgreek }
\newunicodechar{ω}{\textomega}
\newunicodechar{ɛ}{\textepsilon}
\newunicodechar{ɔ}{\textopeno}
\newunicodechar{β}{\textbeta}
\newunicodechar{ɲ}{\textltailn}
\newunicodechar{ɣ}{\textgamma}

\title{A Very Low Resource Language Speech Corpus for Computational Language Documentation Experiments
}

\name{P. Godard$^{\ast}$, G. Adda$^{\ast}$, M. Adda-Decker$^{\dagger}$, J. Benjumea$^{\ddagger}$, L. Besacier $^{\star}$, J. Cooper-Leavitt$^{\ast}$, \\
\bf \large G-N. Kouarata$^{\dagger}$, L. Lamel$^{\ast}$, H. Maynard$^{\ast}$, M. M{\"u}ller$^{\diamond}$, A. Rialland$^{\dagger}$, S. St{\"u}ker$^{\diamond}$, F. Yvon$^{\ast}$, \\
\bf \large M. Zanon-Boito $^{\star}$}

\address{$^{\ast}$LIMSI, CNRS, Université Paris-Saclay, Orsay, France \\
         $^{\dagger}$LPP, CNRS, Paris, France\\
         $^{\ddagger}$ENS, EHESS, PSL Research University, CNRS, INRIA, Paris, France\\
          $^{\star}$LIG, UGA, G-INP, CNRS, INRIA, Grenoble, France\\
          $^{\diamond}$KIT, Karlsruhe, Germany\\
         contact: pierre.godard@limsi.fr, laurent.besacier@univ-grenoble-alpes.fr}

\abstract{
Most speech and language technologies are trained with massive amounts of speech and text information. However, most of the world languages do not have such resources and some even lack a stable orthography. Building systems under these almost zero resource conditions is not only promising for speech technology but also for computational language documentation. The goal of computational language documentation is to help field linguists to (semi-)automatically analyze and annotate audio recordings of endangered, unwritten languages. Example tasks are automatic phoneme discovery or lexicon discovery from the speech signal.
This paper presents a speech corpus collected during a realistic language documentation process. It is made up of 5k speech utterances in {\embosi} (Bantu C25) aligned to French text translations. Speech transcriptions are also made available: they correspond to a non-standard graphemic form close to the language phonology. 
We detail how the data was collected, cleaned and processed and we illustrate its use through a zero-resource task: spoken term discovery. The dataset is made available to the community  for reproducible computational language documentation experiments and their evaluation.
\\ \newline  }

\begin{document}

\maketitleabstract

\section{Introduction}

Many languages will face extinction in the coming decades. Half of the 7,000 languages spoken worldwide are expected to disappear by the end of this century \cite{cambridge-book2011}, and there are too few field linguists to document all of these endangered languages. Innovative speech data collection methodologies \cite{bird_al_2014,blachon2016} as well as computational assistance \cite{addaBulbSLTU2016,stukerBulbCCURL2016} were recently proposed to help them in their documentation and description work. 

%
As more and more researches are related to computational language documentation \cite{duong2016attentional,franke2016phoneme,godard2016preliminary,anastasopoulos2017case}, there is a need of realistic corpora to fuel reproducible and replicable language studies at the phonetic, lexical and syntactic levels. To our knowledge, very few corpora are available for computational analysis of endangered languages.\footnote{We are only aware of a Griko-Italian corpus \cite{grikodatabase}, and of a Basaa-French corpus \cite{hamlaoui2018}.}

Our work follows this objective and presents a speech dataset collected following a real language documentation scenario. It is multilingual ({\embosi} speech aligned to French text) and contains linguists' transcriptions in {\embosi} (in the form of a non-standard graphemic form close to the language phonology). The corpus is also enriched with automatic forced-alignment between speech and  transcriptions. The dataset is made available to the research community\footnote{It will be made available for free from ELRA,
but its current version is online on: \url{https://github.com/besacier/mboshi-french-parallel-corpus}}.
This dataset is part of a larger data collection conducted on {\embosi} language and presented in a companion paper \cite{rialland2018}.

Expected impact of this work is the evaluation of efficient and reproducible computational language documentation approaches in order to face fast and inflexible extinction of languages.  

This paper is organized as follows: after presenting the language of interest (\embosi) in section \ref{mboshi}, we describe the data collection and processing in sections \ref{DC} and \ref{DP} respectively. Section \ref{wordseg} illustrates  its first use for an unsupervised word discovery task. Our spoken term detection pipeline is also presented and evaluated in this section. Finally, section \ref{concl} concludes this work and gives some perspectives


\section{Language Description}
\label{mboshi}

{\embosi} (Bantu C25) is a typical Bantu language spoken in Congo-Brazzaville. It is one of the languages documented by the BULB (Breaking the Unwritten Language Barrier) project~\cite{addaBulbSLTU2016,stukerBulbCCURL2016}.

\paragraph{Phonetics, phonology and transcription}
{\embosi} has a seven vowel system (i, e, ɛ, a, ɔ, o, u) with an opposition between long and short vowels. Its consonantal system includes the following phonemes: p, t, k, b, d, β, l, r, m, n, ɲ, mb, nd, ndz, ng, mbv, f, s, ɣ , pf, bv, ts, dz, w, j. It has a set of prenasalized consonants (mb, nd, ndz, ng, mbv) which are common in Bantu languages \cite{EmbangaAborobongui13,Kouarata2014}.

While the language can be considered as rarely written, linguists have nonetheless defined a non-standard graphemic form for it, considered to be close to the language phonology. Affricates and prenasalized plosives were coded using multiple symbols (e.g.\ two symbols for dz, three for mbv). Long and short vowels were coded respectively as V and as VV.


{\embosi} displays a complex set of phonological rules. The deletion of a vowel before another vowel in particular, common in Bantu languages, occurs at 40\% of word junctions \cite{RiallandEtAl15}. This tends to obscure word segmentation and introduces an additional challenge for automatic processing.

\paragraph{Morphology}

{\embosi} words are typically composed of roots and affixes, and almost always include at least one prefix, while the presence of several prefixes and one suffix is also very common. The suffix structure tends to consist of a single vowel V (e.g. -a or -i) whereas the prefix structure may be both CV and V. Most common syllable structures are V and CV, although CCV may arise due to the presence of affricates and prenasalized plosives mentioned above.

The noun class prefix system, another typical feature of Bantu languages, has an unusual rule of deletion targeting the consonant of prefixes\footnote{A prefix consonant drops if the root begins with a consonant \cite{RiallandEtAl15}.}. The structure of the verbs, also characteristic of Bantu languages, follows: Subject Marker --- Tense/Mood Marker --- Root-derivative Extensions --- Final Vowel. A verb can be very short or quite long, depending of the markers involved.  

\paragraph{Prosody}
{\embosi} prosodic system involves two tones and an intonational organization without downdrift \cite{Rialland16how}. The high tone is coded using an acute accent on the vowel while low tone vowel has no special marker. Word root, prefix and suffix all bear specific tones which tend to be realized as such in their surface forms.\footnote{The distinction between high and low tones is phonological (see \cite{Rialland16how}).} Tonal modifications may also arise from vowel deletion at word boundaries.

A productive combination of tonal contours in words can also take place due to the preceding and appended affixes. These tone combinations play an important grammatical role particularly in the differentiation of tenses. However, in {\embosi}, the tones of the roots are not modified due to conjugations, unlike in many other Bantu languages.

\section{Data Collection}
\label{DC}


We have recently introduced \ligaikuma\footnote{\url{http://lig-aikuma.imag.fr}}, a mobile app specifically dedicated to fieldwork language documentation, which works both on android powered smartphones and tablets \cite{blachon2016}. It relies on an initial smartphone application developed by \cite{bird_al_2014} for the purpose of language documentation with an aim of long-term interpretability. 
We extended the initial app with the concept of retranslation (to produce oral translations of the initial recorded material) and speech elicitation from text or images (to collect speech utterances aligned to text or images). In that way, human annotation labels can be replaced by weaker signals in the form of parallel multimodal information (images or text in another language). {\ligaikuma} also implements the concept of \textit{respeaking}  initially introduced in \cite{woodbury_2003}. It involves listening to an original recording and repeating what was heard carefully and slowly. This results in a secondary recording that is much easier to analyze later on (analysis by a linguist or by a machine).
So far, {\ligaikuma} was used to collect data in three unwritten African Bantu languages in close collaboration with three major European language documentation groups (LPP, LLACAN in France; ZAS in Germany). 

Focusing on {\embosi} data, our corpus was built both from translated reference sentences for oral language documentation \cite{Bouquiaux76} and from a {\embosi} dictionary \cite{Beapami00}. Speech elicitation from text was performed by three speakers in Congo-Brazzaville and led to more than 5k spoken utterances. The corpus is split in two parts (\textit{train} and \textit{dev}) for which we give basic statistics in Table~\ref{table:basic-stats}. We shuffled the corpus prior to splitting in order to have comparable distributions in terms of speakers and origin.\footnote{Either \cite{Bouquiaux76} or \cite{Beapami00}.} There is no text overlap for {\embosi} transcriptions between the two parts.

\begin{table}
  \centering
  \small
  \begin{tabular}{lcrrr}  
  \toprule
  language  & split  & \#sent & \#tokens & \#types \\ \midrule
  Mboshi    & train & 4,616 & 27,563    & 6,196 \\
            & dev   & 514   & 2,993     & 1,146 \\ \midrule
  French    & train & 4,616 & 38,843    & 4,927 \\
            & dev   & 514   & 4,283     & 1,175 \\
  \bottomrule
  \end{tabular}
  \caption{Corpus statistics for the \embosi{} corpus}
  \label{table:basic-stats}
\end{table}

\section{Data Processing}
\label{DP}

	\subsection{Cleaning, Pre-/Post-Processing}
	
	All the characters of the {\embosi} transcription have been checked, in order to avoid multiple encodings of the
	same character. Some characters have also been transcoded so that a character with a diacritic effectively corresponds to the UTF-8 composition of the individual character with the diacritic.
	Incorrect sequences of tones (for instance tone on a consonant) have been corrected. It was also decided to remove the elision symbol in {\embosi}.
	
	On the French side, the translations were double-checked semi-automatically (using linux $aspell$ command followed by a manual process -- 3.3\% of initial sentences were corrected this way). The French translations were finally tokenized (using $tokenizer.perl$ from the Moses toolkit\footnote{\url{http://www.statmt.org/moses/}}) and lowercased. We propose an example of a sentence pair from our corpus in Figure~\ref{fig:example}.


	
      
\begin{figure*}[t]
      \centering
      \small
      \begin{tabular}{lcl} \midrule
      Mboshi & & \texttt{wáá ngá iwé léekundá ngá sá oyoá lendúma saa m ótéma} \\
      French & & \texttt{si je meurs  enterrez-moi dans la forêt oyoa avec une guitare sur la poitrine} \\
     \midrule
      \end{tabular}
      \caption{A tokenized and lowercased sentence pair example in our Mboshi-French corpus.}
      \label{fig:example}
\end{figure*}

	\subsection{Forced Alignment}

  As the linguists' transcriptions did not contain any timing information, the creation of timed alignments was necessary.
  The word and phoneme level alignments were produced with what Cooper-Leavitt et al. refer to as `A light-weight ASR tool' \cite{ltc17}. The alignment tool is an ASR system that is used in a forced-alignment mode. That is, it is used to associate words in the provided orthographic level transcription with the corresponding audio segments making use of a pronunciation lexicon which represents each word with one or more pronunciations (phonemic forms). 
  The word-position-independent GMM-HMM monophone models are trained using the STK tools at LIMSI \cite{HCL15}. A set of 68 phonemes are used to represent the pronunciation dictionary,
  with multiple symbols for each vowel representing different tones~\cite{IS17jcl,BPa1996} and a symbol for silence. The acoustic model is estimated iteratively, with 5 rounds of segmentation and parameter estimation, and the model resulting from the last iteration was used to resegment the data.
  Since vowel elision and morpheme deletion are known to be characteristic of the {\embosi} language, variants explicitly allowing their presence or absence are included in the pronunciation lexicon. Details of how this was implemented can be found in \cite{ltc17}.
  
  
  

\section{Illustration: Unsupervised Word Discovery from Speech}
\label{wordseg}

In this section, we illustrate how our corpus can be used to evaluate unsupervised discovery of word units from raw speech. This task corresponds to \textit{Track 2} in the \textit{Zero Resource Speech Challenge 2017}.\footnote{\url{http://zerospeech.com}} We present below the evaluation metrics used as well as a monolingual baseline system which does not take advantage yet of the French translations aligned to the speech utterances (bilingual approaches may be also evaluated with this dataset but we leave that for future work). 

	\subsection{Evaluation Metrics}
	Word discovery is evaluated using the \textit{Boundary}, \textit{Token} and \textit{Type} metrics from the \textit{Zero Resource Challenge 2017}, extensively described in \cite{ludusan2014} and \cite{zrc2017}. They measure the quality of a word segmentation and the discovered boundaries with respect to the gold corpus. For these metrics, the precision, recall and F-score are computed; the \textit{Token} recall is defined as the probability for a gold word token to belong to a cluster of discovered tokens (averaging over all the gold tokens), while the \textit{Token} precision represents the probability that a discovered token will match a gold token. The F-score is the harmonic mean between these two scores. 
	Similar definitions are applied to the \textit{Type} and \textit{Boundary} metrics.

 
 	\subsection{Unsupervised Word Discovery Pipeline}

Our baseline for word discovery from speech involves the cascading of the following two modules:
\begin{itemize}
    \item unsupervised phone discovery (UPD) from speech: find pseudo-phone units from the spoken input,
    \item unsupervised word discovery (UWD): find lexical units from the sequence of pseudo-phone units found in the previous step.
\end{itemize}

 
          \subsubsection*{Unsupervised phone discovery from speech (UPD)}
          In order to discover a set of phone-like units, we use the KIT system which is a three step process.
          First, phoneme boundaries are found using BLSTM as described in  \cite{franke2016}.
          For each speech segment, articulatory features are extracted (see \cite{essv2017mueller} for more details).
          Finally, segments are clustered based on these articulatory features \cite{mueller2017icassp}. So, the number of clusters (pseudo phones) can be controlled during this process.

\subsubsection*{Unsupervised word discovery (UWD)}

To perform unsupervised word discovery, we use $dpseg$ \cite{Goldwater06contextual,Goldwater09bayesian}.\footnote{\url{http://homepages.inf.ed.ac.uk/sgwater/resources.html}} It implements a Bayesian non-parametric approach, where (pseudo)-morphs are generated by a bigram model over a non-finite inventory, through the use of a Dirichlet-Process. Estimation is performed through Gibbs sampling.

\newcite{godard2016preliminary} compare this method to more complex models on a smaller {\embosi} corpus, and demonstrate that it produces stable and competitive results, although it tends to oversegment the input, achieving very high recall and a lower precision.


\subsection{Results}




Word discovery results are given in Tables~\ref{table:pipeline1}, \ref{table:pipeline2} and \ref{table:pipeline3} for \textit{Boundary}, \textit{Token} and \textit{Type} metrics respectively.

We compare our results to a pure speech based baseline which does pair-matching using locally sensitive hashing applied to PLP features and then groups pairs using graph clustering  \cite{aren}. The parameters of this baseline spoken term discovery system are the same\footnote{Notably the DTW  threshold -- which controls the number of spoken clusters found -- is set to 0.90 in our experiment} as the ones used for the \textit{Zero Resource Challenge 2017} \cite{zrc2017}.
  
A topline where $dpseg$ is applied to the gold forced alignments (phone boundaries are considered to be perfect) is also evaluated.
  
For the pipeline, we experience with different granularities of the UPD system (5, 30 and 60 units obtained after the clustering step). For each granularity, we perform 3 runs and report the averaged results.

We note that the baseline provided by the system of \cite{aren} has a low coverage (few matches). Given that our proposed pipeline provides an exhaustive parse of the speech signals, it is guaranteed to have full coverage and, thus, shows better performance according to the \textit{Boundary} metric. The quality of segmentation, in terms of tokens and types is, however, still low for all systems.  Increasing the number of pseudo phone units improves \textit{Boundary} recall but reduces precision. For \textit{Token} and \textit{Type} metrics, a coarser granularity provides slightly better results.






  \begin{table}[t]
      \centering
      \small
      \begin{tabular}{lcccc} \toprule
      method & \hphantom{3pt} & P & R & F \\ \midrule
      gold FA phones + \dpseg{}	              && 53.8	& 83.5	& 65.4 \\
      \midrule
      \cite{aren}	          && 31.9	& 13.8	& 19.3 \\
      UPD+\dpseg{} pipeline (5 units)	              && 27.4	& 46.5	& 34.5 \\
      UPD+\dpseg{} pipeline (30 units)	              && 24.6	& 58.9	& 34.7 \\
      UPD+\dpseg{} pipeline (60 units)	              && 24.4	& 60.2	& 34.8 \\
     \bottomrule
      \end{tabular}
      \caption{Precision, Recall and F-measure on word boundaries. Pipeline compared with an unsupervised system based on (Jansen and Van Durme, 2011).
      }
      \label{table:pipeline1}
    \end{table}
    
      \begin{table}[t]
      \centering
      \small
      \begin{tabular}{lcccc} \toprule
      method & \hphantom{3pt} & P & R & F \\ \midrule
      gold FA phones + \dpseg{}	              && 20.8	& 35.2	& 26.2 \\
      \midrule
      \cite{aren}	          && 4.5	& 1.6	& 2.4 \\
      UPD+\dpseg{} pipeline (5 units)	              && 2.1	& 4.4	& 2.9 \\
      UPD+\dpseg{} pipeline (30 units)	              && 1.4	& 4.6	& 2.2 \\
      UPD+\dpseg{} pipeline (60 units)	              && 1.4	& 4.7	& 2.1 \\
     \bottomrule
      \end{tabular}
      \caption{Precision, Recall and F-measure on word tokens. Pipeline compared with an unsupervised system based on (Jansen and Van Durme, 2011). 
      }
      \label{table:pipeline2}
    \end{table}

      \begin{table}[t]
      \centering
      \small
      \begin{tabular}{lcccc} \toprule
      method & \hphantom{3pt} & P & R & F \\ \midrule
      gold FA phones + \dpseg{}	              && 7.5	& 13.9	& 9.7 \\
      \midrule
      \cite{aren}	          && 4.6	& 4.8	& 4.7 \\
      UPD+\dpseg{} pipeline (5 units)	              && 2.5	& 6.7	& 3.6 \\
      UPD+\dpseg{} pipeline (30 units)	              && 2.0	& 4.5	& 2.8 \\
      UPD+\dpseg{} pipeline (60 units)	              && 2.0	& 4.4	& 2.7 \\
     \bottomrule
      \end{tabular}
      \caption{Precision, Recall and F-measure on word types. Pipeline compared with an unsupervised system based on (Jansen and Van Durme, 2011).
      }
      \label{table:pipeline3}
    \end{table}

\section{Conclusion}
We have presented a speech corpus in {\embosi} made available to the research community for reproducible computational language documentation experiments. As an illustration, we presented the first unsupervised word discovery (UWD) experiments applied to a truly unwritten language ({\embosi}).


The results obtained with our pipeline on pseudo-phones are still far behind those obtained with gold labels, which indicates that there is room for improvement for the UPD module of our pipeline. The UWD module should, in future work, make use of the bilingual data available to improve the quality of the segmentation.

Future work also includes enriching our dataset with some alignments at the word level, in order to evaluate a bilingual lexicon discovery task. This is possible with encoder-decoder approaches, as shown in \cite{zanon2017}.

As we distribute this corpus, we hope that this will help the community to strengthen its effort to improve the technologies currently available to support linguists in documenting endangered languages.



\label{concl}	
	
\section{Acknowledgements}  
This work was partly funded by the French ANR and the German DFG under grant ANR-14-CE35-0002.

Some experiments reported of this paper were performed during the Jelinek Summer Workshop on Speech and Language Technology (JSALT) 2017 in CMU, Pittsburgh, which was supported by JHU and CMU (via grants from Google, Microsoft, Amazon, Facebook, and Apple). 

\section{Bibliographical References}
\label{main:ref}

\bibliographystyle{lrec}
\bibliography{bulb}


\end{document}